\documentclass{IEEEtran}

\usepackage{graphicx}
\usepackage{epstopdf}
\usepackage{amsmath,amssymb,booktabs,tabularx}
\usepackage[hyphens]{url}
\usepackage{hyperref}
\usepackage{mathrsfs}
\usepackage{float}
\usepackage{multirow}
\usepackage{soul}
\usepackage[table]{xcolor}
\usepackage{comment}
\usepackage{subcaption}
\usepackage{mwe}
\usepackage{slashbox}
\usepackage{authblk}
\usepackage{multicol}
\usepackage{import}
\usepackage{siunitx}
\usepackage{mathtools}

\usepackage{makecell} 

\newcommand{\eg}{e.g.~}

\begin{document}

\title{Synthetic Glacier SAR Image Generation from Arbitrary Masks Using Pix2Pix Algorithm}

\author{\IEEEauthorblockN{
		Rosanna Dietrich-Sussner\textsuperscript{1*}, 
		Amirabbas Davari\textsuperscript{1*}, 
		Thorsten Seehaus\textsuperscript{2}, 
		Matthias Braun\textsuperscript{2},
		Vincent Christlein\textsuperscript{1}, 
		Andreas Maier\textsuperscript{1},
		Christian Riess\textsuperscript{1} \\
	}
	\textsuperscript{1} Department of Computer Science, Friendrich-Alexander University Erlangen-Nürnberg, Germany\\ 
	\textsuperscript{2} Department of Geography \& Geosciences, Friedrich-Alexander University Erlangen-Nürnberg, Germany\\ 
	* Rosanna Dietrich-Sussner and Amirabbas Davari contributed equally to this work.
}

\maketitle

\begin{abstract}
%
%
Supervised machine learning requires a large amount of labeled data to achieve
proper test results. However, generating accurately labeled segmentation maps
on remote sensing imagery, including images from synthetic aperture radar
(SAR), is tedious and highly subjective.
In this work, we propose to alleviate the issue of limited training data by generating synthetic SAR images with the pix2pix algorithm \cite{isola2017image}.
This algorithm uses conditional Generative Adversarial Networks (cGANs) to generate an artificial image while preserving the structure of the input. In our case, the input is a segmentation mask, from which a corresponding synthetic SAR image is generated.
We present different models, perform a comparative study and demonstrate that
this approach synthesizes convincing glaciers in SAR images with promising
qualitative and quantitative results.

\end{abstract}


\section{Introduction}\label{sec:introduction}


\IEEEPARstart{G}{laciers} and ice sheets are undergoing significant changes and
are frequently addressed as good climate indicators.
The position of the calving front of marine or lake terminating glaciers is an
important measure, since recession can lead to destabilization of the glacier
system and subsequent ice mass losses~\cite{furst2016safety}.
Due to the remote location of the majority of the world's ice masses and the spread over large areas, remote sensing data is ideal to monitor ongoing changes.
However, one particular monitoring challenge are sea/lake ice and icebergs,
which often form an ``ice-melange'' covering the water surface in front of the
calving front. The surface texture of the ice-melange and the glacier are often
quite similar, which makes the segmentation difficult.

Most studies of calving front positions manually annotate the front positions,
which is laborious and subjective~\cite{paul2013accuracy}.  Moreover, it does
not scale well to the increasing amount of available remote sensing imagery,
which additionally raises the need for automated analysis methods.
Previous works proposed (semi-)automatic approaches to detect the calving
fronts using techniques like edge detection or image
classification~\cite{baumhoer2018remote}.
Several further studies have explored the potential of CNNs for calving front
mapping. Mohajerani~\emph{et al.} applied a U-Net CNN on multi-spectral Landsat
imagery to detect calving fronts of glaciers in
Greenland~\cite{mohajerani2019detection}. Based on SAR imagery,
Baumhoer~\emph{et al.}~\cite{baumhoer2019automated} and Zhang~\emph{et
al.}~\cite{zhang2019automatically} also proposed U-Net CNNs for mapping calving
fronts.

The performance bottleneck of such supervised learning algorithms is the amount
of training data. Accurately annotated datasets are scarce and expensive to
generate.  Generative Adversarial Networks (GANs) can generate high-quality
samples from a given data distribution~\cite{goodfellow2014generative}. 
Simple GANs learn a plausible embedding of the data without connection to any
labels.  However, conditional Generative Adversarial Networks (cGANs) sample
from a distribution that is conditioned on the
labels~\cite{goodfellow2014generative}. 
Isola~\emph{et al.} propose a cGAN for image-to-image translation, which they
call pix2pix~\cite{isola2017image}.
Their method has been demonstrated on a wide range of tasks such as converting
daytime images to night scenes, black and white photographs to color images,
and Google map view to street view images. 

There exist various works on SAR-to-optical image synthesis using cGANs.
Turnes~\emph{et al.} build on pix2pix algorithm to translate SAR images to optical images introducing atrous convolution~\cite{turnes2020atrous}. Merkle~\emph{et al.} investigate SAR and optical image matching based on the pix2pix idea~\cite{merkle2017possibility}.
Liu~\emph{et al.} generate synthetic SAR imagery from the MSTAR database~\cite{liu2018generating}. Their conditional input is simulated data, but they do not use data paired with annotations.

To our knowledge, the proposed work is the first to generate SAR images from
segmentation masks. The major contribution for our community is the generation
of artificial SAR images with corresponding ground truth segmentation masks.
The benefits of this approach are two-fold: first, the synthesized data can be
used as training data to develop more accurate SAR glacier segmentation models.
Second, to increase the diversity of the training and evaluation data, it
allows to synthesize new shapes and forms of glaciers that are not available in
the hand-annotated datasets.
The remainder of this paper is organized as follows.
In Sec. \ref{sec:pipeline} we describes the pix2pix algorithm and our workflow.
Section \ref{sec:experimental_setup} presents the used dataset and our
experimental setup, followed by qualitative and quantitative results.
Finally, Sec.~\ref{sec:conclusion} provides the conclusions and perspectives for
future work.

\section{Methodology}\label{sec:pipeline}
\begin{figure}[tb]
	\begin{minipage}[b]{0.99\linewidth}
		\centering
		\centerline{\includegraphics[width=1\linewidth]{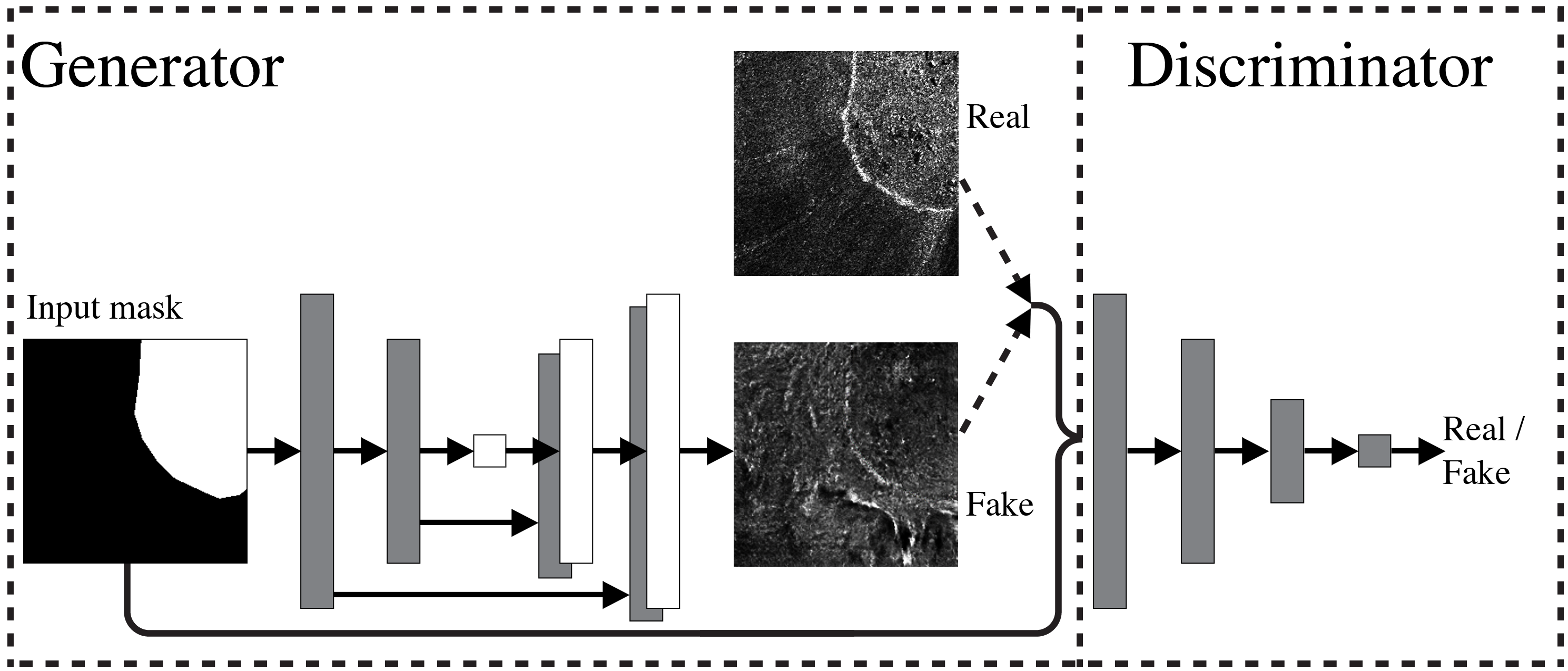}}
	\end{minipage}

	\caption{Network architecture: generator aims to fool the discriminator, which distinguishes real and fake images.}
	\label{fig:architecture}
\end{figure}

\subsection{The Pix2pix conditional GAN}
A GAN consists of two components, namely a generator $G$ and a discriminator $D$.
The generator maps a noise vector $z$ to an output image $y$, $G:z \rightarrow y$. Simultaneously, the discriminator learns to distinguish generated ``fake'' images  from ``real'' images.
$G$ and $D$ are simultaneously optimized with the objective function
\begin{equation}
    \min_{G} \max_{D}  \mathbb{E}_y [\log{D(y)}] + \mathbb{E}_{z}[\log{(1-D(G(z))}]\enspace.
\end{equation}

The conditional GAN extends this idea. It maps a random noise vector $z$ and an observed conditional image $x$ to the output image $y$, $G:(x,z) \rightarrow y$. 
Isola~\emph{et al.}~\cite{isola2017image} propose to combine a cGan objective
\begin{align}
    \mathcal{L}_{cGAN}(G,D) = {} & \mathbb{E}_{x,y}[\log D(x,y)] + \\
   & \mathbb{E}_{x,z}[log(1-D(x,G(x,z))]
\end{align}
with an $L_1$ distance
\begin{equation}\label{eq:l1loss}
    \mathcal{L}_{L_1}(G) = \mathbb{E}_{x,y,z}[ \left\lVert y-G(x,z)\right\rVert_1 ] 
\end{equation}
to obtain a generated image close to the original.
$\mathcal{L}_{L_1}(G)$ only affects the loss of the generator. The resulting
objective function is a minimax game between $G$ and $D$ plus the weighted
$L_1$ term
\begin{equation}\label{eq:pix2pixobjective}
    G^* = \arg \min_{G} \max_{D} \mathcal{L}_{cGAN}(G,D) + \lambda \mathcal{L}_{L1}(G)\enspace.
\end{equation}
%
The generator is modeled as a U-Net, an encoder-decoder network with skip connections~\cite{ronneberger2015u}. The discriminator is a standard classifier called ``PatchGAN''~\cite{isola2017image}. Here, each output position denotes possibility that an $N \times N$ input region is real or fake.

\subsection{Training Pipeline and Parameter Variants}

The proposed workflow is shown in Fig.~\ref{fig:architecture}. In the training phase, generator and discriminator are fed with the preprocessed SAR image patches along with their corresponding segmentation masks. Preprocessing is further described in Sec.~\ref{sec:dataset}.
In each training iteration, $D$ and $G$ are updated according to the objective function in Eq.~\ref{eq:pix2pixobjective}. The discriminator learns to classify pairs of image and mask as real or fake. The generator simultaneously learns to generate images from the input masks that appear increasingly ``real'' to the discriminator.

Besides the original pix2pix model, we explored different parameter variants.
We investigate different kernel sizes in the generator and discriminator as
receptors for spatial information. Another crucial application-dependent
parameter is the weighting $\lambda$ of the two loss functions. We investigated
three different weightings in our comparative analysis.

\section{Evaluation and Experimental Setup}\label{sec:experimental_setup}

\subsection{Dataset}\label{sec:dataset}

\begin{figure}[tb]
	\begin{minipage}[b]{1\linewidth}
		\centering
		\centerline{\includegraphics[width=1\linewidth]{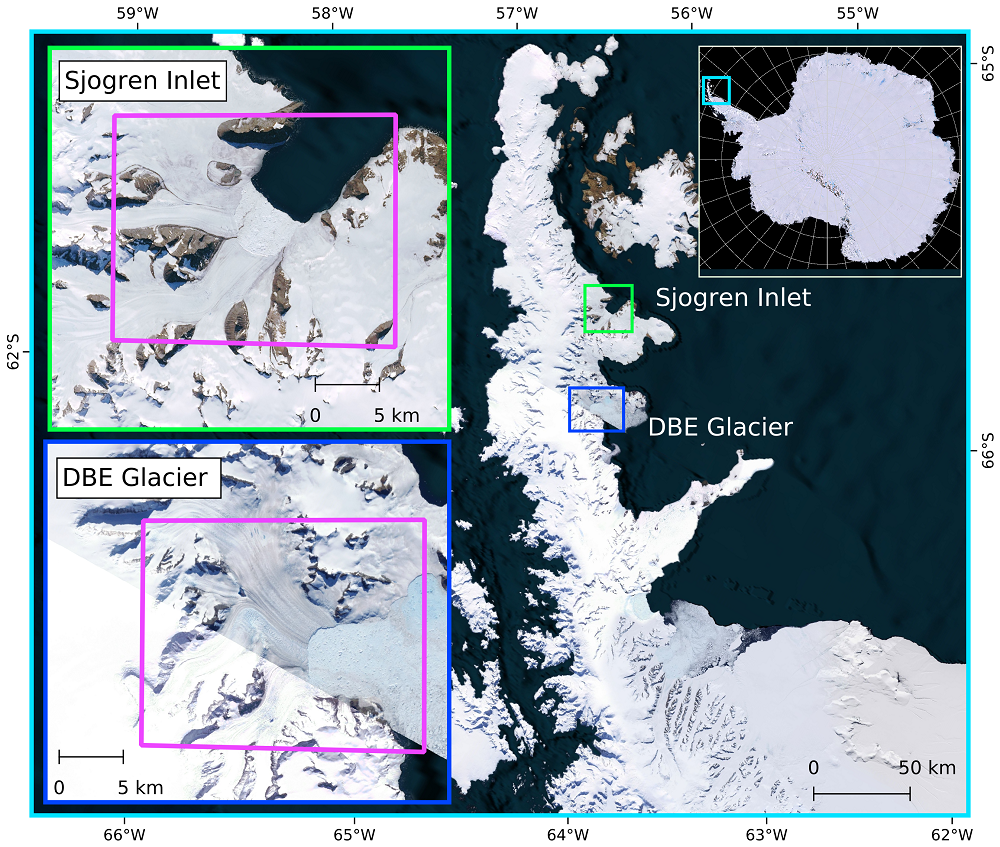}}
	\end{minipage}

	\caption{Map of study sites. Green and blue indicate the analysis regions in the glacier systems. Magenta indicates areas of interest used for the CNN analysis. Background: Bing Aerial Maps © Microsoft; Map of Antarctica: Landsat LIMA Mosaic © USGS, NASA, BAS, NFS.}
	\label{fig:glacierLocation}
\end{figure}

\subsubsection{Study sites}
The Antarctic Peninsula is a hot spot of climate change. The outlet glaciers, typically located in narrow fjords, have undergone significant changes within the last decade.
We selected the Sj{\"o}gren-Inlet (SI) and Dinsmoore-Bombardier-Edgworth (DBE) glacier systems for our analysis. Both were major tributaries to the Prince-Gustav-Channel and Larsen-A ice shelves, which disintegrated in 1995. 

\subsubsection{SAR imagery processing} \label{sec:Preprocessing}
We analyze SAR data from the ERS-1/2, Envisat, RadarSAT-1, ALOS, TerraSAR-X (TSX) and TanDEM-X (TDX) missions, acquired between 1995 and 2014. The imagery is first multi-looked to reduce speckle noise. Based on the ASTER digital elevation model by Cook~\emph{et al.}~\cite{cook2012new}, the intensity imagery is geocoded and orthorectified.

\subsubsection{Training data generation}
The manually mapped calving fronts from Seehaus~\emph{et al.} are used to
generate the training labels~\cite{seehaus2015changes, seehaus2016dynamic}.
The preprocessed SAR acquisitions are cropped to the areas of interest (AOI),
as depicted in Fig.~\ref{fig:glacierLocation}. The ground truth contains two
classes, namely glacier and rocks, and ocean.
%
We extract non-overlapping patches of $256 \times 256$ pixels.
These patches are randomly split into training and validation sets with a ratio
of $90/10$, resulting in $2003$ training and $223$ validation images.

\begin{figure}[tb]
	\begin{minipage}[b]{1\linewidth}
		\centering
		\includegraphics[width=.8\linewidth]{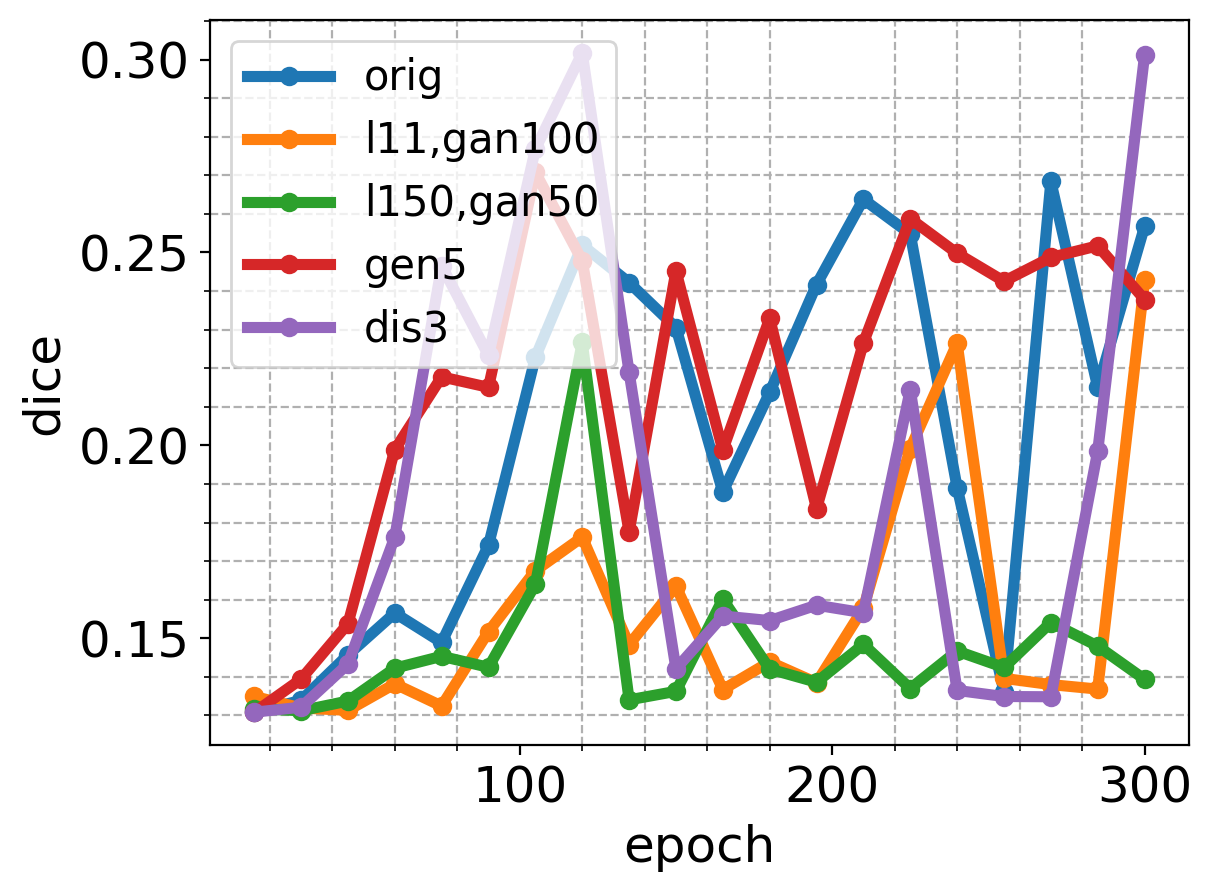}%
	\end{minipage}
	\begin{minipage}[b]{1\linewidth}
		\centering
		\includegraphics[width=.8\linewidth]{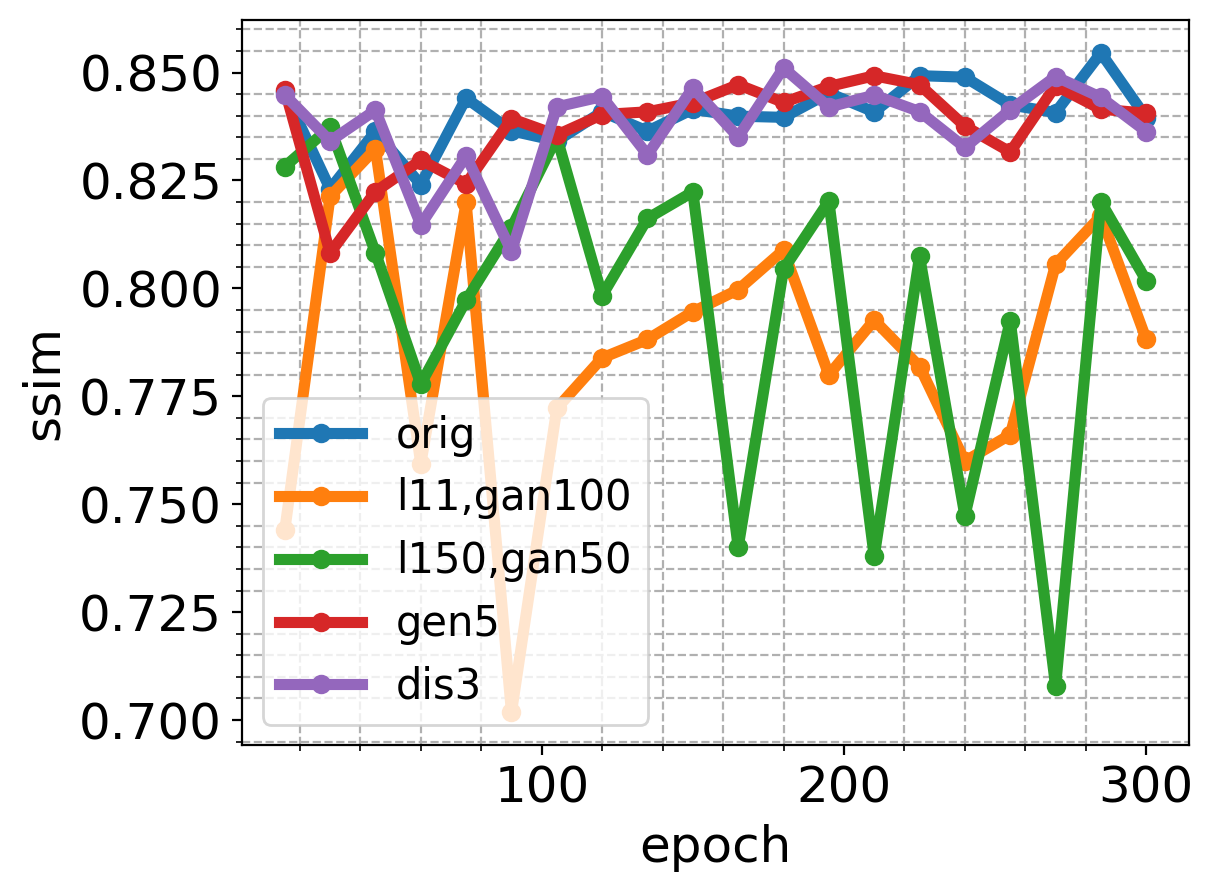}%
	\end{minipage}
	\caption{Quantitative results per training epoch. Top: non-binary dice coefficient. Bottom: structural similarity index measure.}
	\label{fig:dicessim}
\end{figure}

\subsection{Processing Pipeline}
The pix2pix generator consists of $n=16$ layers, in an U-Net architecture, where $8$ layers form the encoder and decoder, respectively, with skip connections between layer $i$ and $n-i$. At each skip connection these layers are concatenated. All encoder layers are of the form Convolution-BatchNorm-LeakyReLU, except the first one which does not apply BatchNorm. The slope of LeakyReLU is $0.2$ and the Convolutions downsample by a factor of $2$, with kernel size $4 \times 4$ and stride $2$.
The first $3$ layers of the decoder consist of Convolution-BatchNorm-Dropout-ReLU layers with a dropout rate of $50\%$. Layers $4$-$7$ are Convolution-BatchNorm-LeakyReLU blocks. Layer $8$ consists of a Convolution and Tanh layer. The convolutions in the decoder upsample by a factor of $2$, with kernel size $4 \times 4$ and stride $2$.

The discriminator is a ``PatchGAN'' with $5$ layers. The first layer 
is a Convolution and LeakyReLU. Then follow $3$
Convolution-BatchNorm-LeakyReLU blocks, and finally a Convolution and Sigmoid
function. Convolutions have kernel sizes of $4 \times 4$, stride $2$, and the
slope of LeakyReLU is $0.2$. Thus,
a $256 \times 256$ input image leads to a $30 \times 30$ output where each
pixel in the discriminator's output represents the possibility of a $70 \times
70$ region in the input to be real or fake. 

We initialize the weights from a Gaussian distribution with mean $0$ and
standard deviation $0.02$. The weights are optimized with Adam with a learning
rate of $0.0002$. For the loss function we use Eq.~\ref{eq:pix2pixobjective}
and weigh the GAN term with $1$ and the $L_1$ term with $100$ in the generator
loss. We refer to this original pix2pix model as \textbf{orig}.

To further investigate possibilities to enhance the pix2pix performance on our SAR dataset, we test the following modifications to the original algorithm:
First, we examine the weight parameters in the generator loss.
We flip the generator loss weighting by multiplying the GAN term with $100$ and the $L_1$ term with $1$, abbreviated as \textbf{l11,gan100}. The goal is to emphasize the GAN term over the $L_1$ distance to the target image, and thereby to
encourage further ``real'' looking SAR images instead of images that merely
copy the training distribution. 
Additionally, we equally weigh both loss terms, abbreviated as \textbf{l150,gan50}, to choose a middle path between both extremes.
Second, we adjust the kernel sizes of the convolution layers. One variant is to
weaken the discriminator by decreasing its kernel size to $3\times 3$,
abbreviated as \textbf{dis3}, which reduces its spatial context. 
We also increase the size of the generator convolution kernels $5 \times 5$,
abbreviated as \textbf{gen5}, to strengthen its spatial context. A drawback
here is the increase in the number of parameters, which leads to longer
training.
%

\subsection{Qualitative and Quantitative Results}\label{sec:results}
Each variant is trained as follows:
Each cGAN is trained for $300$ epochs. Every $15$ epochs, the generator model is saved as a checkpoint. 
In the evaluation procedure, every saved generator model is tested on the $223$ validation masks. Here, each output image is compared to the respective ground truth SAR image.
For a quantitative comparison to the ground truth, we use the non-binary dice coefficient
\begin{equation}\label{eq:dice_nonbin}
    \mbox{Dice}_{\mbox{non-binary}}(X,Y) = \dfrac{2\sum\limits_{i=1}^M X(i)Y(i)}{\sum\limits_{i=1}^M X(i) + \sum\limits_{i=1}^M Y(i)}
\end{equation}
and structural similarity index measure (SSIM)~\cite{wang2004image}. 
The SSIM compares the structure of the output without requiring it to be identical to the target, allowing the generator to produce plausibly looking outputs that are not exact copies of the input.
%
%

Figure~\ref{fig:dicessim} shows the averaged non-binary dice coefficient (top) and SSIM (bottom) for all training epochs and models.
For both metrics, \textbf{orig}, \textbf{gen5} and \textbf{dis3} outperform the other two setups. The dice coefficient oscillates more than SSIM as it measures the absolute pixel-wise similarity of the prediction and the ground truth. Table~\ref{tab:dice_ssim} lists average non-binary dice and SSIM for all models after training for 210 epochs. Here, \textbf{orig} and \textbf{gen5} perform best.
%
%
\begin{table}
\centering
\caption{Averaged non-binary dice coefficient and SSIM at epoch $210$. The models \textbf{orig} and \textbf{gen5} perform best.}
\label{tab:dice_ssim}
\begin{tabular}{ccc}
\hline
model               & non-binary dice & SSIM            \\ \hline
\textbf{orig}       & \textbf{0.2638} & 0.8409          \\
\textbf{gen5}       & 0.2266          & \textbf{0.8492} \\
\textbf{dis3}       & 0.1566          & 0.8448          \\
\textbf{l11,gan100} & 0.1578          & 0.7926          \\
\textbf{l150,gan50} & 0.1484          & 0.7379          \\ \hline
\end{tabular}
\end{table}

\begin{figure*}[tb]
	\begin{minipage}[b]{0.139\linewidth}
		\centering
		\centerline{\includegraphics[width=1\linewidth]{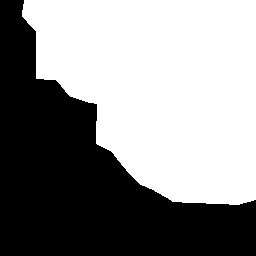}}
	\end{minipage}
	\begin{minipage}[b]{0.139\linewidth}
		\centering
		\centerline{\includegraphics[width=1\linewidth]{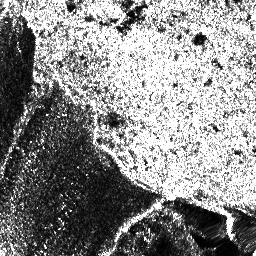}}
	\end{minipage}
	\begin{minipage}[b]{0.139\linewidth}
		\centering
		\centerline{\includegraphics[width=1\linewidth]{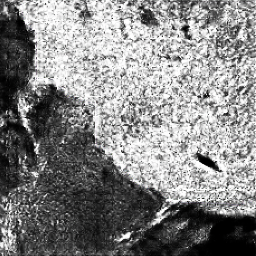}}
	\end{minipage}
	\begin{minipage}[b]{0.139\linewidth}
		\centering
		\centerline{\includegraphics[width=1\linewidth]{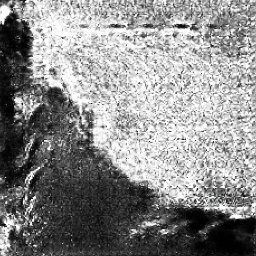}}
	\end{minipage}
	\begin{minipage}[b]{0.139\linewidth}
		\centering
		\centerline{\includegraphics[width=1\linewidth]{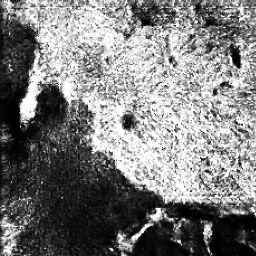}}
	\end{minipage}
	\begin{minipage}[b]{0.139\linewidth}
		\centering
		\centerline{\includegraphics[width=1\linewidth]{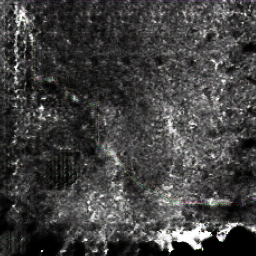}}
	\end{minipage}
	\begin{minipage}[b]{0.139\linewidth}
		\centering
		\centerline{\includegraphics[width=1\linewidth]{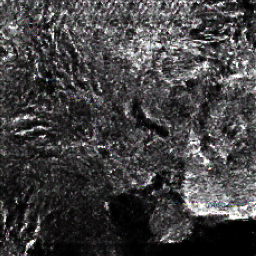}}
	\end{minipage}
	\vspace{-0.3cm}
	
	\begin{minipage}[b]{0.139\linewidth}
		\centering
		\centerline{\includegraphics[width=1\linewidth]{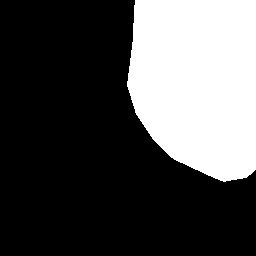}}
		\centerline{(a)}
	\end{minipage}
	\begin{minipage}[b]{0.139\linewidth}
		\centering
		\centerline{\includegraphics[width=1\linewidth]{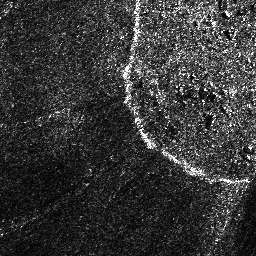}}
		\centerline{(b)}
	\end{minipage}
	\begin{minipage}[b]{0.139\linewidth}
		\centering
		\centerline{\includegraphics[width=1\linewidth]{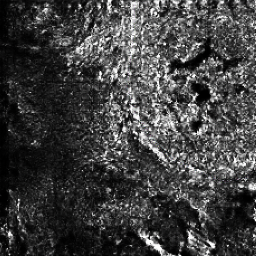}}
		\centerline{(c)}
	\end{minipage}
	\begin{minipage}[b]{0.139\linewidth}
		\centering
		\centerline{\includegraphics[width=1\linewidth]{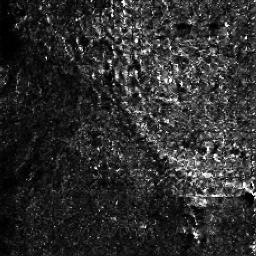}}
		\centerline{(d)}
	\end{minipage}
	\begin{minipage}[b]{0.139\linewidth}
		\centering
		\centerline{\includegraphics[width=1\linewidth]{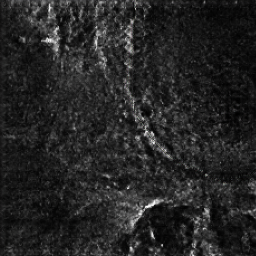}}
		\centerline{(e)}
	\end{minipage}
	\begin{minipage}[b]{0.139\linewidth}
		\centering
		\centerline{\includegraphics[width=1\linewidth]{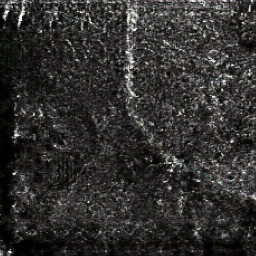}}
		\centerline{(f)}
	\end{minipage}
	\begin{minipage}[b]{0.139\linewidth}
		\centering
		\centerline{\includegraphics[width=1\linewidth]{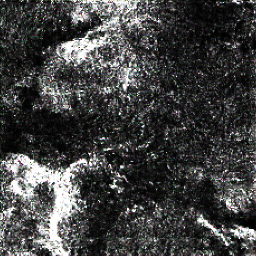}}
		\centerline{(g)}
	\end{minipage}
	\caption{Results after 210 epochs: (a) input mask, (b) target image, (c) \textbf{orig}, (d) \textbf{dis3}, (e) \textbf{gen5}, (f) \textbf{l11,gan100} (g) \textbf{l150,gan50}.}
	\label{fig:comp_all_300}
\end{figure*}

Figure~\ref{fig:comp_all_300} shows a qualitative comparison of two example patches after $210$ training epochs. From left to right, it shows the (a) input mask, (b) target image, and results for (c) \textbf{orig}, (d) \textbf{dis3}, (e) \textbf{gen5}, (f) \textbf{l11,gan100}, and (g) \textbf{l150,gan50}. 
In both examples, the last two variants \textbf{l11,gan100} and
\textbf{l150,gan50} are quite dissimilar to the target image and their
structure barely resembles the input mask. This observation is in agreement
with the quantitative results in Fig.~\ref{fig:dicessim}. Hence, we note that
the weighting modifications of these variants are not helpful.

%
The three other models \textbf{orig}, \textbf{dis3}, and \textbf{gen5} exhibit
better performance. 
On top, all three models generate visually convincing outputs, except
\textbf{dis3} in (d) that exhibits an unnatural repetitive structure in the
upper region.  On bottom, the results are less similar to the target. All
models preserve the structure of the input mask. However, \textbf{orig} in (c)
generates a very noisy image, and \textbf{dis3} in (d) the glacier's boundary
can barely be recognized. \textbf{Gen5} in (e) performs marginally better here.
The boundary is more defined and the structure appears most natural.

\section{Conclusions and Future Work}\label{sec:conclusion}
Limited training data is a common issue in supervised learning on remote
sensing imagery. In this work, we propose to compensate this limitation by
generating artificial SAR images of glacier calving fronts via the pix2pix
algorithm.  We investigate different parametrization of the weighting terms
and kernel sizes. The best results are achieved with the original pix2pix model
and when improving the generator by increasing its kernel size. 

We hope that these preliminary investigations inspire further optimizations towards a fully automated generation of large-scale training datasets for glacier calving fronts. For example, different objective functions or a modified layer structure can be explored. Most importantly, it will be interesting to thoroughly evaluate the benefit of this additional synthetic data to improve the training of existing segmentation frameworks. 
Also, it is possible to generate images from artificial masks. Thereby, unknown
shapes of glaciers can be examined by geography experts. Further, more
channels, \eg elevation, could be added to the plain SAR image to add further
application domains.


\bibliographystyle{ieeetr}
\bibliography{References}

\begin{thebibliography}{10}

\bibitem{isola2017image}
P.~Isola, J.-Y. Zhu, T.~Zhou, and A.~A. Efros, ``Image-to-image translation
  with conditional adversarial networks,'' in {\em Proceedings of the IEEE
  conference on computer vision and pattern recognition}, pp.~1125--1134, 2017.

\bibitem{furst2016safety}
J.~J. F{\"u}rst, G.~Durand, F.~Gillet-Chaulet, L.~Tavard, M.~Rankl, M.~Braun,
  and O.~Gagliardini, ``The safety band of antarctic ice shelves,'' {\em Nature
  Climate Change}, vol.~6, no.~5, pp.~479--482, 2016.

\bibitem{paul2013accuracy}
F.~Paul, N.~E. Barrand, S.~Baumann, E.~Berthier, T.~Bolch, K.~Casey, H.~Frey,
  S.~Joshi, V.~Konovalov, R.~Le~Bris, {\em et~al.}, ``On the accuracy of
  glacier outlines derived from remote-sensing data,'' {\em Annals of
  Glaciology}, vol.~54, no.~63, pp.~171--182, 2013.

\bibitem{baumhoer2018remote}
C.~A. Baumhoer, A.~J. Dietz, S.~Dech, and C.~Kuenzer, ``Remote sensing of
  antarctic glacier and ice-shelf front dynamics—a review,'' {\em Remote
  Sensing}, vol.~10, no.~9, p.~1445, 2018.

\bibitem{mohajerani2019detection}
Y.~Mohajerani, M.~Wood, I.~Velicogna, and E.~Rignot, ``Detection of glacier
  calving margins with convolutional neural networks: A case study,'' {\em
  Remote Sensing}, vol.~11, no.~1, p.~74, 2019.

\bibitem{baumhoer2019automated}
C.~A. Baumhoer, A.~J. Dietz, C.~Kneisel, and C.~Kuenzer, ``Automated extraction
  of antarctic glacier and ice shelf fronts from sentinel-1 imagery using deep
  learning,'' {\em Remote Sensing}, vol.~11, no.~21, p.~2529, 2019.

\bibitem{zhang2019automatically}
E.~Zhang, L.~Liu, and L.~Huang, ``Automatically delineating the calving front
  of jakobshavn isbr{\ae} from multitemporal terrasar-x images: a deep learning
  approach,'' {\em The Cryosphere}, vol.~13, no.~6, pp.~1729--1741, 2019.

\bibitem{goodfellow2014generative}
I.~Goodfellow, J.~Pouget-Abadie, M.~Mirza, B.~Xu, D.~Warde-Farley, S.~Ozair,
  A.~Courville, and Y.~Bengio, ``Generative adversarial nets,'' in {\em
  Advances in neural information processing systems}, pp.~2672--2680, 2014.

\bibitem{turnes2020atrous}
J.~N. Turnes, J.~D.~B. Castro, D.~L. Torres, P.~J.~S. Vega, R.~Q. Feitosa, and
  P.~N. Happ, ``Atrous cgan for sar to optical image translation,'' {\em IEEE
  Geoscience and Remote Sensing Letters}, 2020.

\bibitem{merkle2017possibility}
N.~Merkle, P.~Fischer, S.~Auer, and R.~M{\"u}ller, ``On the possibility of
  conditional adversarial networks for multi-sensor image matching,'' in {\em
  2017 IEEE International Geoscience and Remote Sensing Symposium (IGARSS)},
  pp.~2633--2636, IEEE, 2017.

\bibitem{liu2018generating}
W.~Liu, Y.~Zhao, M.~Liu, L.~Dong, X.~Liu, and M.~Hui, ``Generating simulated
  sar images using generative adversarial network,'' in {\em Applications of
  Digital Image Processing XLI}, vol.~10752, p.~1075205, International Society
  for Optics and Photonics, 2018.

\bibitem{ronneberger2015u}
O.~Ronneberger, P.~Fischer, and T.~Brox, ``U-net: Convolutional networks for
  biomedical image segmentation,'' in {\em International Conference on Medical
  image computing and computer-assisted intervention}, pp.~234--241, Springer,
  2015.

\bibitem{cook2012new}
A.~J. Cook, T.~Murray, A.~Luckman, D.~G. Vaughan, and N.~E. Barrand, ``A new
  100-m digital elevation model of the antarctic peninsula derived from aster
  global dem: methods and accuracy assessment.,'' {\em Earth system science
  data.}, vol.~4, no.~1, pp.~129--142, 2012.

\bibitem{seehaus2015changes}
T.~Seehaus, S.~Marinsek, V.~Helm, P.~Skvarca, and M.~Braun, ``Changes in ice
  dynamics, elevation and mass discharge of dinsmoor--bombardier--edgeworth
  glacier system, antarctic peninsula,'' {\em Earth and Planetary Science
  Letters}, vol.~427, pp.~125--135, 2015.

\bibitem{seehaus2016dynamic}
T.~C. Seehaus, S.~Marinsek, P.~Skvarca, J.~M. van Wessem, C.~H. Reijmer, J.~L.
  Seco, and M.~H. Braun, ``Dynamic response of sj{\"o}gren inlet glaciers,
  antarctic peninsula, to ice shelf breakup derived from multi-mission remote
  sensing time series,'' {\em Frontiers in Earth Science}, vol.~4, p.~66, 2016.

\bibitem{wang2004image}
Z.~Wang, A.~C. Bovik, H.~R. Sheikh, and E.~P. Simoncelli, ``Image quality
  assessment: from error visibility to structural similarity,'' {\em IEEE
  transactions on image processing}, vol.~13, no.~4, pp.~600--612, 2004.

\end{thebibliography}

\end{document}